%% file: main.tex
\def\year{2020}\relax
\documentclass[letterpaper]{article} 
\usepackage{aaai20}  
\usepackage{times}  
\usepackage{helvet} 
\usepackage{courier}  
\usepackage[hyphens]{url}  
\usepackage{graphicx} 
\usepackage{graphicx}  
\usepackage{amsfonts}       
\usepackage{graphicx}
\usepackage{amssymb}

\usepackage{tikz}
\usetikzlibrary{arrows.meta}
\usepackage{forest}
\usepackage{bm}
\usepackage{pgfplots}
\pgfplotsset{compat=1.14}

\newcommand{\citet}[1]{\citeauthor{#1}~\shortcite{#1}}
\newcommand{\citep}{\cite}

\newcommand{\todo}[1]{}

\title{Graph Representations for Higher-Order \\ Logic and Theorem Proving}

\author{
  Aditya Paliwal\thanks{Google AI Resident} \\
  Google Research \\
  \texttt{adipal@google.com} \\
  \And
  Sarah Loos\\
  Google Research \\
  \texttt{smloos@google.com} \\
  \And
  Markus Rabe\\
  Google Research \\
  \texttt{mrabe@google.com} \\
  \AND
  Kshitij Bansal\\
  Google Research \\
  \texttt{kbk@google.com} \\
  \And
  Christian Szegedy\\
  Google Research \\
  \texttt{szegedy@google.com} \\
}

\begin{document}

\maketitle

\begin{abstract}
This paper presents the first use of graph neural networks (GNNs) for higher-order proof search and demonstrates that GNNs can improve upon state-of-the-art results in this domain. Interactive, higher-order theorem provers allow for the formalization of most mathematical theories and have been shown to pose a significant challenge for deep learning. Higher-order logic is highly expressive and, even though it is well-structured with a clearly defined grammar and semantics, there still remains no well-established method to convert formulas into graph-based representations. In this paper, we consider several graphical representations of higher-order logic and evaluate them against the HOList benchmark for higher-order theorem proving.
\end{abstract}

\input{intro}

\input{related_work}
\input{graph_representations}
\input{graph_nets}
\input{experiments}
\input{conclusion}

\bibliographystyle{aaai}
\bibliography{paper}

\input{statistics}

\end{document}

%% file: intro.tex
\section{Introduction}
\label{sec:intro}

Mathematics poses a particularly attractive learning challenge, as it can be seen as a test-bed for general-purpose reasoning.
HOList~\citep{bansal2019holist} is a recently published learning environment for mathematics consisting of a stateless theorem proving API, a benchmark consisting of over twenty thousand mathematical theorems and their proofs, and a neural theorem prover called DeepHOL.
It builds on HOL Light~\citep{Harrison96}, an interactive theorem prover that has been used to formalize several mathematical theories, including topology, multivariate calculus, real and complex analysis, geometric algebra, measure theory and the Kepler conjecture~\citep{hales2017formal}.
The HOList environment and benchmark allows us to measure progress in automated mathematical reasoning, in particular for machine learning techniques.

The percentage of theorems that can be proven automatically by the neural theorem prover DeepHOL presented by~\citet{bansal2019holist} appears to be comparable with state-of-the-art algorithms in higher-order reasoning, which typically build on advanced backtracking search algorithms and special-purpose proof calculi~\citep{kaliszyk2014learning,bentkamp2018superpositionforlambdafreehigherorderlogic}.
While these results were very promising, the reference models presented by~\cite{bansal2019holist} are still relatively naive---in fact we show that we can beat their best models with a bag-of-words model.
The quest for model architectures that show non-trivial understanding of higher-order logic is thus wide-open.

In this work, we explore the use of the tree structure (e.g. the abstract syntax tree) of logic expressions for learning representations.
Most earlier works in this area used TreeRNNs with the idea that the embedding of each node needs to summarize the semantics of the subexpression.
However, TreeRNNs (and likewise sequence models such as LSTMs) have not shown strong performance gains over baselines on logical reasoning tasks.
We believe this is because TreeRNNs fail to consider the \emph{context} of subexpressions---when computing the embedding of an internal node in the syntax tree, TreeRNNs only consider the embeddings of its child-nodes, but never the parent.

In this paper, we consider the syntax trees of formulas as graphs, where each node has edges to its children and also to its parent(s), and apply message-passing graph neural networks (GNNs)~\citep{scarselli2009graph,li2015gated,GilmerSRVD17,gnn_survey_2019} instead of TreeRNNs.
We focus on imitation learning, i.e. learning from human proofs.
The proofs we learn from in the HOList dataset were written by human mathematicians interacting with a computer-based theorem prover.
To ensure that we make meaningful progress, we evaluate all our models by measuring how many theorems they manage to prove when integrated with the DeepHOL neural theorem prover.
In Section~\ref{sec:experiments} we find that GNNs significantly improve performance, and achieve state-of-the-art performance for higher-order logic proof search by a wide margin.
Our best model automatically proves nearly $50\%$ of theorems in the validation set.

We present and compare several graph representations for higher-order logic (Section~\ref{sec:graph-representations}) and suggest a simple, message-passing, GNN architecture (Section~\ref{sec:gnn}).
We demonstrate that minor changes in graph representations can have significant effects on the performance of message passing GNNs.
We additionally confirm our hypothesis that the context of subexpressions is crucial, by first restricting GNNs to pass messages only upwards in the syntax tree (similar to TreeRNNs), and, second, to pass messages only downwards in the syntax tree.
The experiment clearly shows that the second approach, which emphasizes the context of expressions, outperforms the first.

%% file: related_work.tex
\section{Related Work}
\label{sec:related-work}

Graph Neural Network (GNN) is an umbrella term for several kinds of neural networks that operate on graph-structured data \citep{gnn_survey_2019,GilmerSRVD17,xu2019powerful}. This family of neural networks has been successfully applied in several domains ranging from computer vision \citep{RaposoSBPLB17,SantoroRBMPBL17}, to predicting properties of molecules \citep{GilmerSRVD17}, to traffic prediction \citep{LiYSL17}.
\cite{battaglia_survey} show that GNNs are capable of manipulating structured knowledge in the data and hence are a natural choice to learn meaningful representations for higher-order logic.

Similar to our work, \citet{wang2017premise} apply GNNs to higher-order logic; they use graph representations of higher-order logic formulas for premise selection, resulting in a significant improvement in prediction accuracy.
Our work extends their contributions by using GNNs not only for premise selection, but for predicting tactics and tactic arguments at every step of the proof.
While the difference seems minor, it has a big effect on our contributions: it allows us to evaluate our models on a theorem proving benchmark where we demonstrate that our models actually prove more theorems. Because \citet{wang2017premise} only predict the premises at the theorem level, they can not generate each individual proof step, and are therefore not able to use their models to generate proofs; instead they use a proxy metric that measures against existing human proofs. We are able to provide end-to-end metrics on the percentage of theorems from the validation set that can be proved using our models.
This ensures that our models are learning something useful about mathematics instead of learning to exploit syntactic tricks on the data.

Our findings also differ from the results of \citet{wang2017premise} in a crucial aspect: we found that representations that share sub-expressions are significantly stronger than representations that use tree representations.
Our graph representation allows for more sharing between expressions, as we merge variable nodes, even if they belong to different binders.
Also, our representation includes all type information and has a special node for function applications, which allows us to treat variables, abstractions, and quantifiers in a uniform way.

Similar to this work, GamePad \citep{huang2018gamepad}, TacticToe \citep{gauthier2017tactictoe}, CoqGym~\citep{YangDeng/2019/LearningToProveTheoremsViaInteractingWithProofAssistants}, and Proverbot9001~\citep{SanchezSternAlhessiSaulLerner/2019/GeneratingCorrectnessProofsWithNNNs} use imitation learning on human proofs collected from the tactics-based higher-order provers Coq and HOL4.

We elected to use the HOList benchmark as opposed to the GamePad or CoqGym datasets as it spans a large variety of mathematics.
We additionally wanted to ensure that our models performed well when used to guide proof search, which is made simple in HOList.

Traditionally, automated theorem provers were developed for first order logic, as it is easier to create complete and sound provers for it. However, most of the serious human formalization efforts~\citep{gonthier2008formal,gonthier2013odd,hales2017formal} were performed using interactive proof assistants like Coq and HOL Light~\citep{Harrison96}, which are based on higher order logic (in which quantification over arbitrary propositions is allowed).
However, the automation of higher-order theorem proving, especially at the tactic level, is a more recent research area. Deep learning has made inroads into various types of logic reasoning, for example to guide SAT solvers~\citep{selsam2019guiding}, QBF solvers~\citep{Lederman2018QBF}, and for inductive reasoning~\citep{dong2018neural}.

Many authors have used TreeRNN (or TreeLSTM) architectures to encode logical statements based on their abstract syntax tree representations \citep{evans2018can,huang2018gamepad,loos2017deep}. While this seems like a natural way to encode tree-structured data, this approach greatly restricts the flow of information across the tree, especially between siblings. TreeRNNs enforce the property that a subexpression will always have the same embedding, regardless of the context in which the expression appears. While this has computational advantages, as it allows embeddings of subexpressions to be cached, it may also limit the ability of the network to effectively encode semantic information across (structurally) long distances.

Conversely, message-passing graph neural networks do not have a single, static embedding for each subexpression, but rather evolve embeddings for every node based on both its children and parents.  Perhaps more importantly, when the graph representation allows subexpression sharing, these embeddings can draw on context from multiple occurrences.  Our experimental results clearly demonstrate the importance of subexpression sharing.

%% file: graph_representations.tex
\section{Graph Representations of HOL}
\label{sec:graph-representations}

\begin{figure*}[h]
\centering
\input{figures/tree_vs_shared}
\caption{Comparison of two graph representations of $\forall x:\: x=x$. Types (e.g. {\tt fun}, {\tt bool}, and type variable {\tt A}) are printed in orange. Left: AST representation. Right: AST with subexpression sharing (edge labels ensure the left/right order of children is preserved, but are omitted for readability in this figure). Sharing reduces the graph size from 27 nodes to 15; the amount of sharing increases as the terms grow larger.}
\label{fig:sepxression_refl}
\end{figure*}
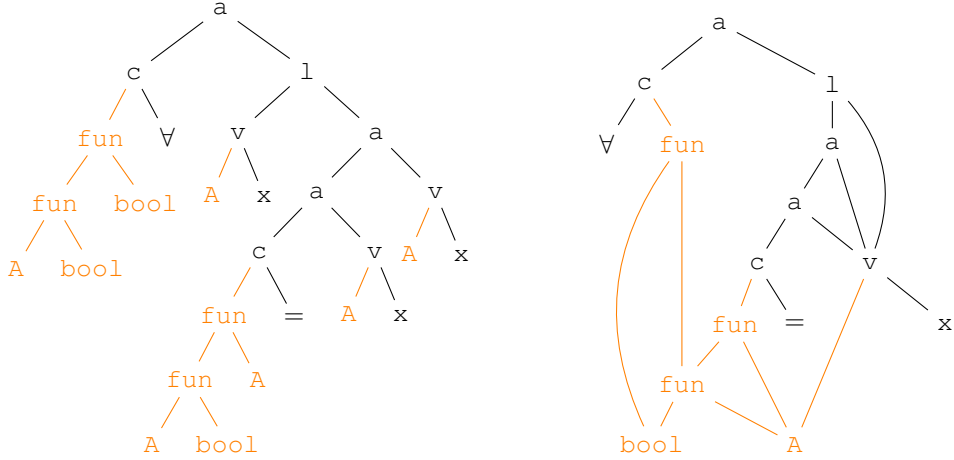

In this section we describe our graph representations of higher-order logic (HOL) terms.
The HOList benchmark provides its data in the form of S-expressions.
For example, the term $f(x)$, which applies a variable $f$ to another variable $x$, is represented as the following string: \texttt{(a (v (fun A B) f) (v A x)))}.
The S-expression breaks the function application down into fundamental elements, which we explain in the following:
The token \texttt{a} indicates a function application, which always has two children; the function to be applied and the argument.
The token \texttt{v} indicates that the function to be applied is a variable.
This variable is then described as having type \texttt{fun A B} (i.e. a function mapping from some type A to some potentially different type B) and having the name \texttt{f}.
Similarly, the expression \texttt{(v A x)} describes the argument of the function application as a variable \texttt{x} with type \texttt{A}.

There are only a small number of these fundamental building blocks of HOList's S-expressions: \texttt{a} for function applications, \texttt{v} for variables, \texttt{l} for lambda expressions (also called abstractions), \texttt{c} for constants, and some additional ones for type constructors such as \texttt{fun}.
All other tokens are names of either variables, constants, or types, of which there are around 1200 in the dataset.
Even quantifiers have no special treatment and are simply represented as functions, using the definition $(\forall) \triangleq \lambda f.\: (f=\lambda x.\: \mathit{True})$.

That is, S-expressions essentially represent the abstract syntax tree of higher-order logic expressions.
We consider the abstract syntax tree (AST) as a graph with directional edges, where each node has edges to its children and also to its parent(s).
Edges are labeled with the index of the child, which makes the original string representation recoverable from the graph.

In addition to the plain AST, we consider several modifications to the graph representations, which we study in Section~\ref{sec:experiments}:
\begin{itemize}
\item We share nodes of syntactically equal subexpressions (\emph{subexpression sharing}).
\item We share equal leaves of the AST (\emph{leaf sharing}).
\item We replace all variable names by \texttt{x} (\emph{variable blinding}).
\item We add random edges (\emph{random}).
\item We remove all edges from nodes to their parents and only keep those to their children (\emph{top down}).
\item We remove all edges from nodes to their children and only keep those to their parents (\emph{bottom up}).
\end{itemize}

\emph{Subexpression sharing} merges all nodes in the AST that represent the same expression.
In Figure~\ref{fig:sepxression_refl} we illustrate the difference between the abstract syntax tree representation with (right) and without (left) subexpression sharing.
The variable \texttt{x} of type \texttt{A}, which occurs three times in the abstract syntax tree, now occurs only once.
It is represented by the S-expression $(v~x~A)$, and the root node of this expression now has three parents that keep track of the locations of each of its original occurrences.
Note that sub-expression sharing also happens over types.
In this example, variable \texttt{x} has type \texttt{A}, so every other expression with type or sub-type \texttt{A} is now connected through this node.
For detailed statistics of how subexpression sharing compresses the data, see Appendix \ref{sec:graph-stats}. 

Leaf sharing is a middle ground between the very verbose AST representation and the aggressive graph reduction that is subexpression sharing.
Instead of merging nodes that represent equal subexpressions, we only merge leaves with the same token, such as \texttt{A}, \texttt{bool}, or \texttt{x}.

\emph{Variable blinding} replaces all variable names by \texttt{x}. This transformation is applied only after the graph is constructed and does not modify the graph structure. Hence, different variables still have different graph nodes and are thus distinguishable - just their names are the same. This allows us to study how much our networks rely on variable names.

Restricting our graphs to only have \emph{top down} edges or only have \emph{bottom up} edges allows us to study the question of how important the context of subexpressions is for their embeddings.
Bottom up loosely resembles TreeRNNs, as it computes its embeddings only from the embeddings of its children.
Top down, on the other hand, disallows nodes to see their children, but allows them to see their context.

Adding random edges to the graph representation allows us to study if additional connectivity helps the neural networks to propagate information through the graphs.
We chose to add 3 outgoing edges per node to random targets, as this approximates the construction of expander graphs, which provides excellent connectivity properties of the resulting graph with little overhead.
We label the random edges to make them distinguishable from regular edges in the graph.

%% file: figures/tree_vs_shared.tex
\begin{forest}
for tree={
  before typesetting nodes={content={\texttt{#1}}}
}
[,phantom,for descendants={l=0mm}
[a,s sep=1mm
  [c
    [fun, color=orange,edge={orange},for descendants={color=orange,edge={orange}},
        [fun
            [A]
            [bool]
        ]
        [bool]
    ]
    [$\forall$]
  ]
  [l
      [v
        [A, color=orange,edge={orange}]
        [x]
      ]
      [a,s sep=0pt
        [a
          [c
            [fun, color=orange,edge={orange},for descendants={color=orange,edge={orange}}
                [fun
                    [A]
                    [bool]
                ]
                [A]
            ]
            [${=}$]   
          ]
          [v
            [A, color=orange,edge={orange}]
            [x]
          ]
        ]
        [v
          [A, color=orange,edge={orange}]
          [x]
        ]
      ]
  ]
]
]
\end{forest}
\qquad\qquad
%
\begin{tikzpicture}
\begin{scope}[every node/.style={font=\ttfamily}]
    \node (root) at (0,0) {a};
    \node (forallconstant) at (-1,-.8) {c};
    \node[color=orange] (foralltype) at (-.5,-1.6) {fun};
    \node[color=orange] (bool) at (-.9,-5.6) {bool};
    \node[color=orange] (typeA) at (1,-5.6) {A};
    \node[color=orange] (typeAtoBool) at (-.5,-4.8) {fun};
    \node (forallname) at (-1.5,-1.6) {$\forall$};
    \node (lambda) at (1.5,-0.8) {l};
    \node (equalityapp) at (1.5,-1.6) {a};
    \node (equalityapp2) at (1,-2.4) {a};
    \node (equalityconst) at (.5,-3.2) {c};
    \node (equalityname) at (1,-4) {{$=$}};
    \node[color=orange] (equalitytype) at (0.2,-4) {fun};
    \node (variablex) at (2,-3.2) {v};
    \node (namex) at (3,-4) {x};
\end{scope}

\begin{scope}[>={Stealth},
              every edge/.style={draw}]
    \path [-] (root) edge (forallconstant);
    \path [-,color=orange] (forallconstant) edge (foralltype);
    \path [-,color=orange] (foralltype) edge (typeAtoBool);
    \path [-,color=orange] (typeAtoBool) edge (typeA);
    \path [-,color=orange] (typeAtoBool) edge (bool);
    \path [-,color=orange] (foralltype) edge[bend right=30] (bool);
    \path [-] (forallconstant) edge (forallname);
    \path [-] (root) edge (lambda);
    \path [-] (lambda) edge[bend left=30] (variablex);
    \path [-] (lambda) edge (equalityapp);
    \path [-] (equalityapp) edge (equalityapp2);
    \path [-] (equalityapp) edge (variablex);
    \path [-] (equalityapp2) edge (variablex);
    \path [-] (equalityapp2) edge (equalityconst);
    \path [-,color=orange] (equalityconst) edge (equalitytype);
    \path [-] (equalityconst) edge (equalityname);
    \path [-,color=orange] (equalitytype) edge (typeAtoBool);
    \path [-,color=orange] (equalitytype) edge (typeA);
    \path [-] (variablex) edge (namex);
    \path [-,color=orange] (variablex) edge (typeA);
\end{scope}
\end{tikzpicture}

%% file: graph_nets.tex
\section{Model Architecture}
\label{sec:model_architectures}

In this section we present our neural network architecture, starting with basics of message-passing GNNs in Section~\ref{sec:gnn}, which we use for embedding statements in higher-order logic.
We then detail the prediction tasks necessary for guiding proof search and our imitation learning approach, which trains on human proofs, in Section~\ref{sec:full_model}.

\subsection{Graph Neural Networks}
\label{sec:gnn}

\begin{figure*}
\centering
  \includegraphics[width=0.9\linewidth, trim=0 15mm 45mm 0, clip]{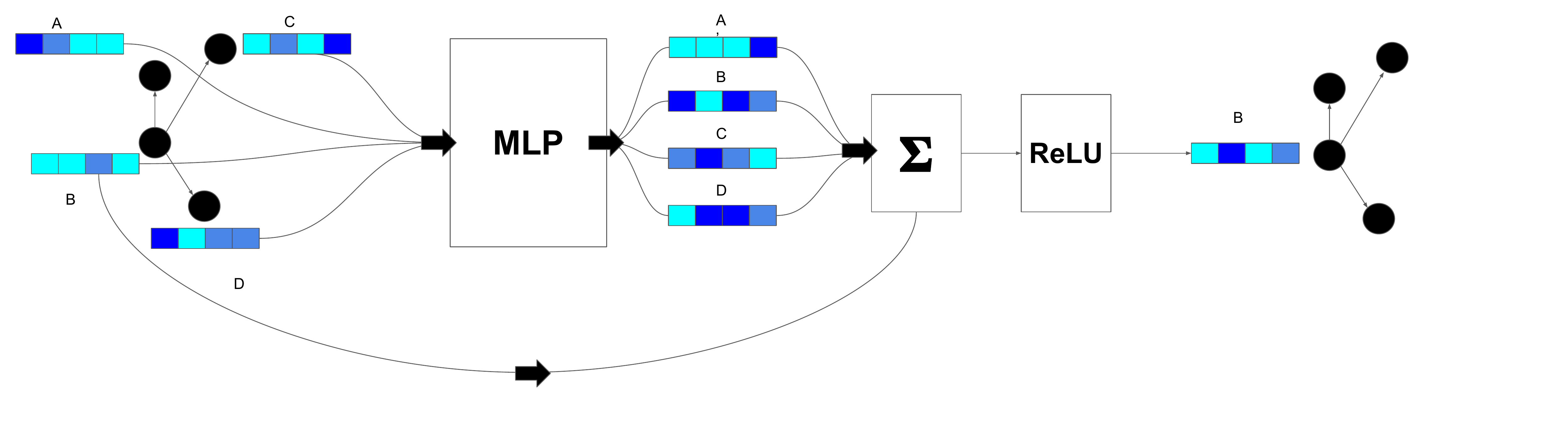}
  \caption{A single Graph Neural Network neighborhood aggregation step for node B.
  The features over the neighborhood are passed through an MLP, followed by a summation and a non-linearity.
  The output that follows is the hidden feature for node B for the next step.}
  \label{fig:gnn-node-aggregation}
\end{figure*}

Graph neural networks (GNNs) compute embeddings for nodes in a graph via consecutive rounds (also called \emph{hops}) of end-to-end differentiable message passing.
The input to a GNN is a labeled graph $G = (V, E, l_V, l_E)$ where $V$ is a set of nodes, $E$ is a set of directed edges, $l_V$ maps nodes to a fixed vocabulary of tokens, and $l_E$ maps each edge $e$ to a single scalar indicating if the edge is to (or from) the first or the second child, encoded as 0 and 1.
Each token $t$ in the vocabulary is associated with a trainable feature vector $\bm x_t$.

The GNN computes node embeddings $\bm{h}_v$ for each node $v \in V$ via an iterative message passing process of $T$ rounds:

    1. Embed nodes $n$ and edges $e$ into high dimensional space using multi-layer perceptrons:
    $$\bm{h}_v^1 = \textrm{MLP}_V(\bm{x}_{l_V(v)})$$
    $$\bm{h}_e = \textrm{MLP}_E(l_E(e))$$

    2. For each round $t \in \{2, \dots, T\}$, and for each edge $(u,v) = e \in E$, pass the node embeddings from the previous step $\bm{h}^{t-1}_u$ and $\bm{h}^{t-1}_v$ and the edge embedding $\bm{h}_e$ into an MLP to generate messages. 
    When computing the messages for a node $v$, we distinguish between messages from parent nodes $\bm s_{u,v}$ and from child nodes $\hat{\bm s}_{u,v}$:
    $$\bm{s}_{u,v}^t = \textrm{MLP}^t_{\textrm{edge}}([\bm{h}_u^{t-1}, \bm{h}_v^{t-1}, \bm{h}_e])$$
    $$\hat{\bm{s}}^{t}_{u,v} = \hat{\textrm{MLP}^t}_{\textrm{edge}}([\bm{h}_u^{t-1}, \bm{h}_v^{t-1}, \bm{h}_e])$$

    3. To summarize the messages for node $v \in V$, we sum over the messages from parents and from children separately before passing them through an MLP and adding them to the previous embedding: 
    $$\bm{h}_v^{t} = \bm{h}_v^{t-1} + \textrm{MLP}_{\textrm{aggr}}\left(\left[\bm{h}_v^{t-1}, \sum\frac{\bm{s}^{t}_{u, v}}{p(v)}, \sum \frac{\hat{\bm{s}}^{t}_{u,v}}{c(v)}\right]\right)$$

$\textrm{MLP}_V$, $\textrm{MLP}_E$, $\textrm{MLP}^t_{\textrm{edge}}$, $\hat{\textrm{MLP}^t}_{\textrm{edge}}$ and $\textrm{MLP}_{\textrm{aggr}}$ are multi-layer perceptrons, $p(v)$ is the number of parents of node $v$, $c(v)$ is the number of children of node $v$ $\left[,\right]$ is the vector concatenation operator, and $\bm{h}_v^t$ represents the embedding of node $v$ after $t$ rounds of message passing. A multi-layer perceptron ($\textrm{MLP}$), is a function $\textrm{MLP}: \mathbb{R}^a \to \mathbb{R}^b$, that maps vectors in the input space to the output space via successive applications of linear transformations and non-linear activations.

The final set of node embeddings returned by the Graph Neural Network is given by $\bm{h}_v = \bm{h}_v^T$ where $T$ is the number of message passing rounds. These node embeddings represent information from the $T$-hop neighborhood of each node in the graph. A single step of message passing over a single node of a graph is shown in Figure \ref{fig:gnn-node-aggregation}.

\subsection{Complete Model Architecture}
\label{sec:full_model}

\begin{figure}[t]
  \includegraphics[width=\linewidth]{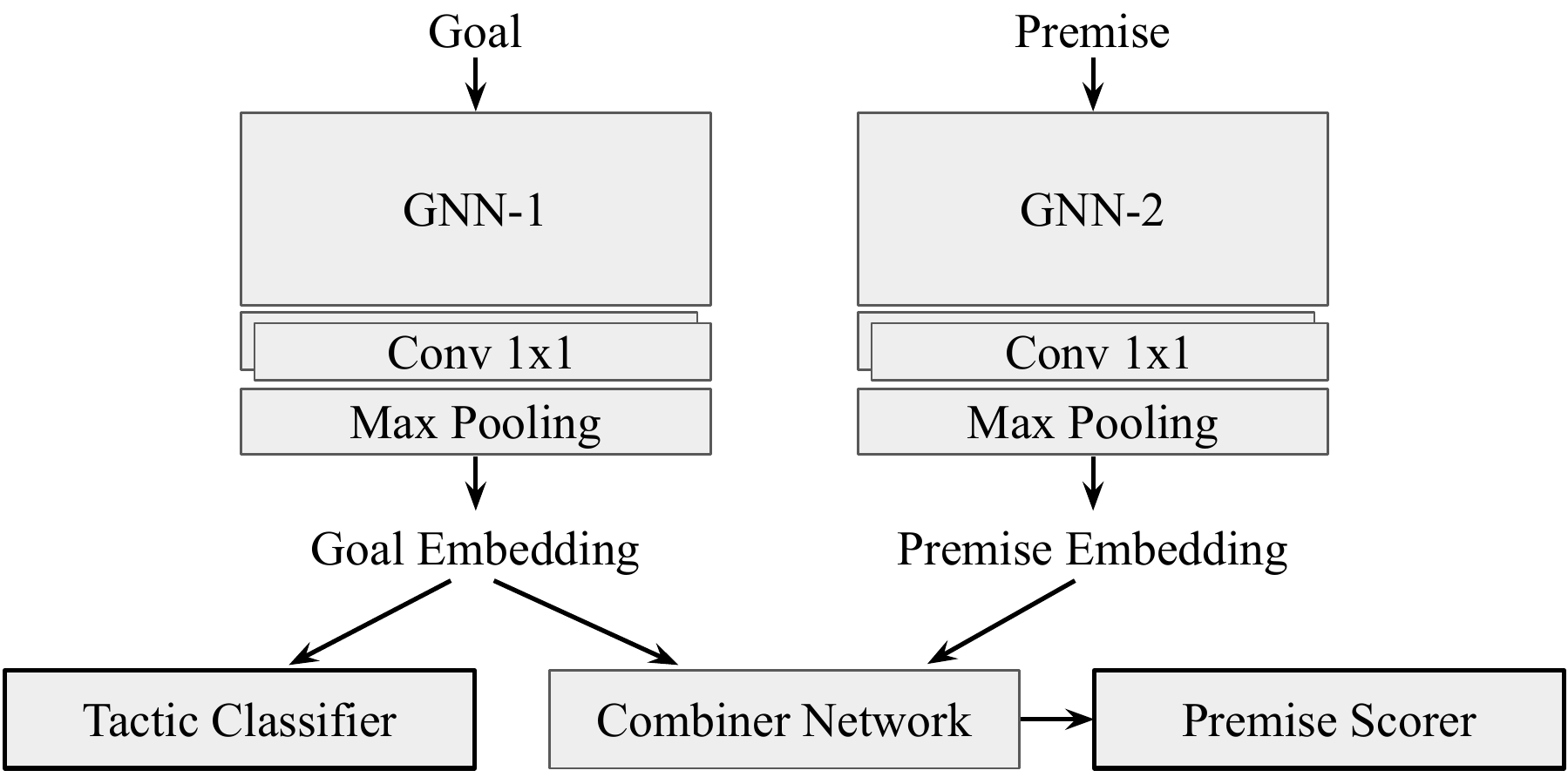}
  \caption{
  Diagram of the model architecture.
  At every step of the proof, the current goal is embedded using GNN-1.
  Based on the goal embedding, the model predicts the next tactic to apply from a fixed set of 41 tactics.
  The goal embeddings are also used to score every preceding theorem, called premises.
  Premises are also embedded using a graph neural network (GNN-2).
  Higher scores indicate the given premise is predicted to be useful for proving the current goal.  Full details presented in Section \ref{sec:full_model}.
  }
  \label{fig:architecture}
\end{figure}

The full architecture is depicted in Figure \ref{fig:architecture}. 
The network takes the current goal (an intermediate state in a proof search) and a candidate premise.

It then generates node embeddings $G(g)$ and $P(p)$ for the graph representations of both the goal $g$ and premise $p$ using identical GNNs; GNN-1 and GNN-2 do not share weights.
To make the architecture computationally tractable, the depth of each node embedding is relatively small (128). 
We do two 1x1 convolutions to expand the depth (to 512 and then to 1,024) immediately before max pooling, resulting in a goal embedding and premise embedding, each of size 1,024. 

The tactic classifier selects from a fixed set of 41 tactics.
It uses two fully connected layers, followed by a linear layer to produce logits for a softmax tactic classifier.
The combiner network concatenates the goal and premise embeddings, as well as their element-wise multiplication, followed by three fully connected layers.
The model uses sigmoid cross-entropy to score how useful an individual premise is for the given goal. It was important to apply dropout throughout the network.

This architecture was trained on the proofs written by humans and released in the HOList dataset.
For evaluation, we plug the models into the breadth-first proof search provided by the HOList environment. We first start with the top level goal which is the top level theorem to be proved. To process a goal (or ``subgoals'' for newly created goals), we apply one of the tactics that may take a list of already proven theorems as parameters.\footnote{The interpretation of the tactic parameters depends on the type of the tactic. For {\tt MESON\_TAC}, these are premises that the tactic can use for proving the statement in HOL Light's built-in first-order reasoning algorithm. For {\tt REWRITE\_TAC}, these should be equations that are applied for rewriting the goal statement.}
Once a goal is set to be processed, we pick the top-$k_1$ highest scoring tactics from the softmax tactic classifier~(see Figure \ref{fig:architecture}). Then the combiner network is evaluated for each $(g, p_i)$ pairs of the current goal $g$ and possible tactic parameter $p_i$ (which includes all the definitions and theorems preceding the top level goal in the theorem database). Note that this computation is accelerated significantly by pre-computing the premise-embeddings $P(p_i)$, and therefore only the combiner network has to be evaluated for each pair of embeddings $G(g), P(p_i))$, where $G(g)$ denotes the goal embedding of the currently processed goal. In total, there are 19,262 theorems and definitions in the theorem database.  Only the top-$k_2$ highest scoring premises are chosen as tactic arguments. In our setting, we used $k_1=5$, $k_2=20$.

A tactic application might fail or may be successful. Failed applications are logged, but ignored for proof search. If the tactic application is successful, it might close (that is: prove) the goal or generate a new list of subgoals (or proof obligations). The original goal is closed if each of its subgoals is closed. During proof search, the HOList proof-search graph maintains several alternative branches of subgoals to be closed. Proof search stops if the top-level goal closes, that is if at least one of the alternative branches closes.

Note that in contrast to earlier premise selection works, we generate new tactic parameter lists for each subgoal during proof-search, not just a single list of premises for the top-level goal.

%% file: experiments.tex
\section{Experiments}
\label{sec:experiments}

Here we present our experimental results on the HOList benchmark~\citep{bansal2019holist},
which is based on the complex analysis corpus of HOL Light~\citep{Harrison96}. The training set is derived from the proofs of 10,200 top-level theorems and has around 375,000 proof steps (subgoals) to learn from. We have evaluated the end-to-end prover performance on the standard validation split of HOList which consists of 3,225 held out theorems of the complex analysis corpus.

In total, there are 19,262 theorems and definitions in the HOList corpus.  These theorems are proved in a sequential order. So, for any given goal, all preceding theorems and definitions in the corpus are eligible premises.  For each tactic application, we select the top 20 premises after ranking the eligible premises with the premise scorer.

\subsection{Evaluation Metrics}

The primary way that we evaluate our models is by using them to guide the proof search in the HOL Light theorem prover, and then counting how many theorems each trained model can prove. However, it takes several hours to run the prover on the validation set, even in a distributed manner.  If we were to use this metric during training, it would slow down training of our models by several orders of magnitude, so we instead train using two proxy metrics: 
\begin{itemize}
    \item We look at the accuracy of the tactic prediction output, which decides which one of the 41 possible tactics is to be applied for a given goal.  It is often the case that more than one choice of tactic would work, but to determine if a tactic is correct, we would need to try it in the prover. Instead, for the proxy metric, we only count a tactic correct if it was the exact tactic applied in the human proof, as our data contains only a single proof per theorem.
    \item We monitor the relative prediction accuracy for the tactic parameter selection, which is the ratio of cases in which a true tactic parameter is scored higher than some randomly sampled parameter.  Again, there is often more than one ``true parameter'', but we only consider the set of premises that were used as tactic parameters in the human proof.
\end{itemize}

\begin{table}
\centering
\caption{Our best model, compared against previous state of the art and bag of words model as baselines. }
    \begin{tabular}{lr}
    \hline
    Network Architecture & \% Proofs Closed \\
    & (Validation Set) \\
    \hline
    Baseline: S-expression as a string & \\
    WaveNet \citep{bansal2019holist} & 32.65\% \\
    Baseline: Bag of words & \\ 
    Max pooling only & 37.98\% \\
    \hline
    {\bf Subexpression sharing} & \\
    {\bf 12-hop GNN} & {\bf 49.95\%} \\
    \hline
\end{tabular}
\label{tab:proofs-closed-baselines}
\end{table}

\begin{table*}[t]
\centering
\caption{Percentage of proofs closed on the validation set. Representations are defined in Section \ref{sec:graph-representations}. Two AST results indicated with an asterisk (*) required a smaller batch size due to memory constraints.}
    \begin{tabular}{lccccc}
    \hline
    {\bf Representation} & {\bf 0 Hops} & {\bf 2 Hops} & {\bf 4 Hops} & {\bf 8 Hops} & {\bf 12 Hops} \\
    \hline
    Abstract syntax trees (AST) & 40.18\% & 43.84\% & 44.58\% & 46.66\%$^*$ & 45.67\%$^*$ \\
    Leaf sharing & 41.76\% & 33.89\% & 29.24\% & 29.51\% & 30.51\% \\
    Leaf sharing + variable blinding & 31.78\% & 32.18\% & 32.80\% & 30.04\% & 31.00\% \\
    {\bf Subexpression sharing} & 40.86\% & 42.94\%& 46.94\%& 47.22\% & {\bf 49.95\%} \\
    Subexpression sharing + variable blinding & 31.75\% & 34.44\% & 35.96\% & 34.07\% & 37.36\% \\
    Subexpression sharing + random & 41.24\% & 43.68\% & 43.84\% & 42.63\% & 42.94\% \\
    Subexpression sharing + top down & 40.55\% & 43.59\% & 45.51\% & 48.24\% & 48.40\% \\
    Subexpression sharing + bottom up & 39.72\% & 40.58\% & 41.16\% & 41.86\% & 40.99\% \\
    \hline
    \end{tabular}
\label{tab:proofs-closed}
\end{table*}

We then perform the computationally expensive evaluation only on the best checkpoint (according to the proxy metrics): running the prover once over the validation set using the parameterized tactics predicted by the checkpoint to guide proof search. 
This {\it percentage of proofs closed} in the validation set is the primary metric that we use to establish the proving power of a given checkpoint.
It evaluates whether a trained model is effective at automatically proving newly encountered theorems.

\subsection{Results and Analysis}

We train several models using the architecture described in Section \ref{sec:full_model}, varying the number of message passing rounds in the GNN and the representation of the higher-order logic formulas as graphs as described in Section \ref{sec:graph-representations}.

We compare our best-performing model against baselines in Table \ref{tab:proofs-closed-baselines}. Our 12-hop shared subexpression GNN achieves the state-of-the-art result on the HOList benchmark, closing 49.95\% of the proofs. This is a major improvement on the baseline models in the HOList paper (32.65\%) and even outperforms the reported 38.9\% of their best reinforcement learning experiment.
We also observe that a simple bag-of-words model with large word embedding sizes also manages to close $37.98\%$ of all proofs. The GNN models with zero message passing steps (0 Hops) are roughly equivalent to the bag of words max pooling model, except that the 0 Hops models also have additional MLPs between the word embedding lookup step and the max pooling step.

In Table~\ref{tab:proofs-closed} we can see that the number of proofs closed increases with additional hops (message passing steps) for ASTs with and without subexpression sharing. This indicates that additional information about the structure of higher-order logic formulas provides the model with the capacity to learn better embeddings.

Variable blinding, which initializes every token representing a variable name to the same (trained) embedding, allows us to evaluate the importance of human-assigned variable names.
Even though constant values are not altered and the expressions remain semantically equivalent after variable blinding, removing variable names causes a significant drop in performance.
This suggests that human-chosen variable names are important and should not be discarded, even when the resulting formula is semantically equivalent.

The graph representation with random edges improves the performance of the 2-hop network, but not for networks with four or more hops. This suggests that the addition of random edges increases the field-of-view into the graph for the networks, but the benefits diminish and ultimately are harmful for networks with more hops.

Subexpression sharing on ASTs results in a directed acyclic graph, so we can experiment with restricting message passing to flow only from children to parents (bottom up), or from parents to children (top down).
While allowing information to flow in both directions results in the best performance, Table~\ref{tab:proofs-closed} shows that top down message passing significantly outperforms bottom up.
This experiment demonstrates that the context of all the places a subexpression occurs in the larger formula may be more important than the value of the subexpression itself.
It's worth noting that this is opposite the bottom up direction of information flow used in TreeRNNs, where information starts at the leaves and then is propagated to the root.
However, our neural networks have the benefit of aggregating over all nodes in the graph after message passing is complete, a step not taken in TreeRNNs.

While leaf sharing had the effect of compressing the tree representations by approximately 40\%, this representation also incurred a large drop in performance.
Most notably, leaf sharing models which used message passing were well below the 0-hop baseline in all cases.
This result demonstrates the huge impact the representation has on the GNN performance.

Analyzing only the best-performing representation (subexpression sharing, indicated in bold), we find that models with more hops not only close more proofs, but also find slightly longer proofs on average (1.93 steps for 12-hops vs 1.86 for 0-hops).
Another indicator that the quality of the choices made by GNNs increases with with the number of hops is that the percentage of successful tactic applications increased from 33.8\% for 0-hops, to 38.6\% for 12-hops.  

\subsection{Implementation Details}
In this section, we describe our implementation and hardware setup, including hyper-parameter settings and a more detailed description of the losses we used. These hyper-parameters did not vary between experiments unless otherwise noted, and are presented here to aid reproducibility. 
\todo{Discuss hyperparameter sweep/give some indication that these params are reasonable.}

\paragraph{Choosing Negative Examples.}
For each theorem in the HOList training set, there is only one human proof.  We generate positive training examples from each step of the human proof with three values: the goal that is being proved at that step, the tactic applied to it, and a list of theorem parameters passed to the tactic (or a special token if no parameters were used).  To generate positive (goal, premise) pairs, we sample a premise from this list of theorem parameters.  In each batch, we sample 16 positive (goal, premise) pairs from our training data.

Because we only have successful human proofs, we select our negative training examples by sampling uniformly from the set of all theorems that were used at least once as positive training examples.  Every goal in the batch is then paired with 15 randomly sampled theorems and labeled as negative examples (adding 15x16=240 negative examples to the batch).

Since the combiner network is quite small relative to the GNN embedding networks, we reuse all 256 embedded premises (positive and negative) as negative premise examples for the remaining goals in the batch (adding 15x256=3,840 negative examples to the batch), giving a total batch size of 4,096. 

\paragraph{Loss Functions.}
In addition to the cross-entropy loss for both tactics and pairwise scores as described in Section \ref{sec:model_architectures}, we also use the AUCROC loss \citep{burges2005learning,eban2016scalable} to directly relate positive and negative premise examples within a batch.
The AUCROC loss minimizes the area under the ROC curve and is implemented as follows.
$$AUCROC_b = \sum_{i} \sum_{j} loss(logit_i - logit_j)$$
$$loss(l) = \ln(1 + e^{-l})$$ 
Where $i$ ranges over the positive premises in batch $b$, and $j$ ranges over the negatives in $b$.  Because our final prediction task is to rank premises for just one given goal, we double the loss for logits that compare positive and negative premises for the same goal.

For the total loss, we take a weighted sum of the cross-entropy loss on the tactic classifier (weight~=~1.0), the cross-entropy loss on the pairwise scorer (weight~=~0.2), and the AUCROC loss (weight~=~4.0).

\paragraph{Optimizers, Learning Rate, Dropout, and Polyak Averaging.}
For training, we use an Adam Optimizer \citep{kingma2014adam} with initial learning rate 0.0001 and decay rate 0.98. Excluding the two GNNs, dropout is added before every dense layer of the network with a ``keep probability'' of $0.7$.
For our evaluation checkpoints, we also use moving exponential parameter averaging with rate 0.9999 per step \citep{polyak1990new,polyak1992acceleration}.

\paragraph{GNN Hyperparameters.}
The token features $\bm{x}_{t}$ are trainable embedding vectors of size 128. The edge labels $l_E(e)$ are non-trainable binary values $\{0, 1\}$ which indicate if the edge connects a left child or the right child.
The graph neural network begins by projecting the node features $\bm{x}_{l_V(v)}$ and edge features $l_E(e)$ to vectors of size 128 using an MLP with two hidden layers of sizes 256 and 128 with ReLU activations ($\textrm{MLP}_V$ and $\textrm{MLP}_E$). The resulting embeddings are denoted by $\bm{h}^1_v$ and $\bm{h}_e$.
We then perform $t$ rounds of message passing as per the equations shown in Section~\ref{sec:gnn}. $\textrm{MLP}_{\textrm{edge}}$, $\hat{\textrm{MLP}}_{\textrm{edge}}$ and $\textrm{MLP}_{\textrm{aggr}}$ have an identical configuration with two layers, with hidden sizes 256 and 128 and ReLU activations. The MLPs do not share weights across message passing steps.
The final node embedding has size 128.
A dropout of 0.5 is applied to all MLPs.

The node embeddings $\bm{h}_v$ returned by the graph neural network are then aggregated into a single vector that represents the embedding of the entire graph.
Using two Conv~$1~\times~1$ layers, we expand the 128 dimensional node embeddings to 512, and then 1024, with ReLU activations and a dropout rate of 0.5. Then we perform max pooling over all node embeddings to create a single vector of size 1024.

\paragraph{Hardware.}
We used eight NVIDIA Tesla V100 GPUs for distributed training, an additional GPU was used purely for evaluation, and we maintained a separate parameter server on a CPU machine.

%% file: conclusion.tex
\section{Conclusions and Future Work}
\label{sec:conclusion}

We present the first use of graph neural networks to guide higher-order theorem proving. We experiment with several graph representations of higher-order logic expressions and demonstrate that graph neural networks can significantly improve over non-structured representations, especially when the structure of the graph allows for sharing of sub-expressions. 
We observed that increasing the number of message passing steps resulted in improved accuracy on the training and evaluation sets, and also led to a larger percentage of closed proofs.
Using GNNs, we are able to significantly improve previous state-of-the-art results for imitation learning on the HOList theorem set and proof search environment.

In the experiments presented in this paper, we predict tactics and their arguments by looking only at the conclusion of the current sub-goal and ignoring any local assumptions that could be crucial to the proof.
This is a serious limitation for our system, and in future work we would like to include the local assumptions list when generating the embedding of the goal.
We expect this to be a natural extension for GNNs, because they can easily extend graph representations to include additional expressions.

%% file: statistics.tex
\clearpage
\appendix
\section{Appendix}

\subsection{Statistics over the Theorem Database}
\label{sec:graph-stats}

We analyzed how much the different graph representations affect the graph size. 
Below we present a histogram of the number of nodes of all the theorems in the HOList theorem database. We can see that the shared subexpression representation leads to a significant reduction in graph sizes. It is noteworthy that it almost eliminates the tail of the distribution.

\includegraphics[width=\linewidth]{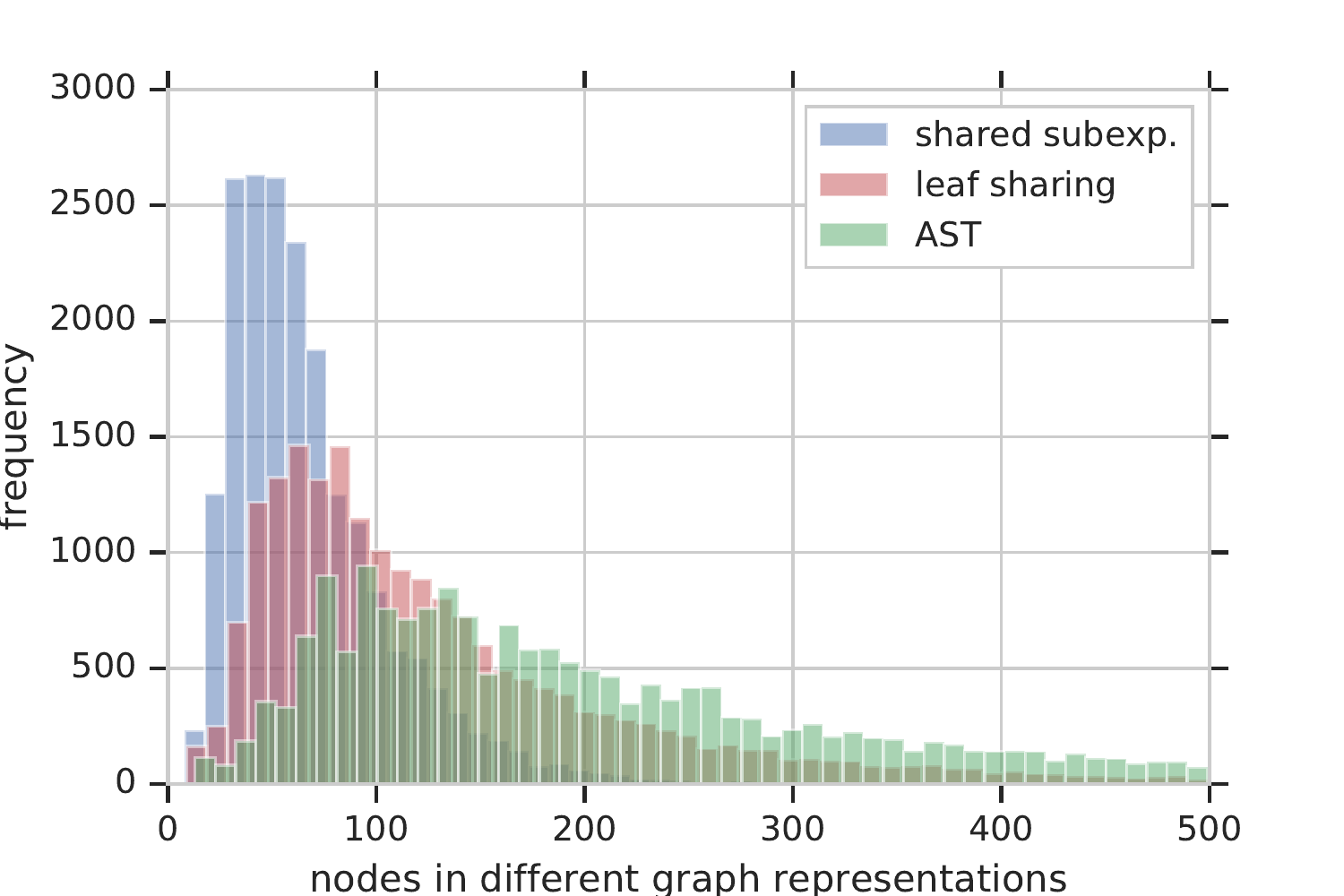}

We also measured the maximal path length from the root node of the graphs, which we call \emph{depth}.
The data indicates that the 12 hops that our best graph neural networks do are not really really sufficient to propagate information from all leafs to the root node, or vice versa.
Models with even more hops might therefore help to increase the performance further.

\includegraphics[width=\linewidth]{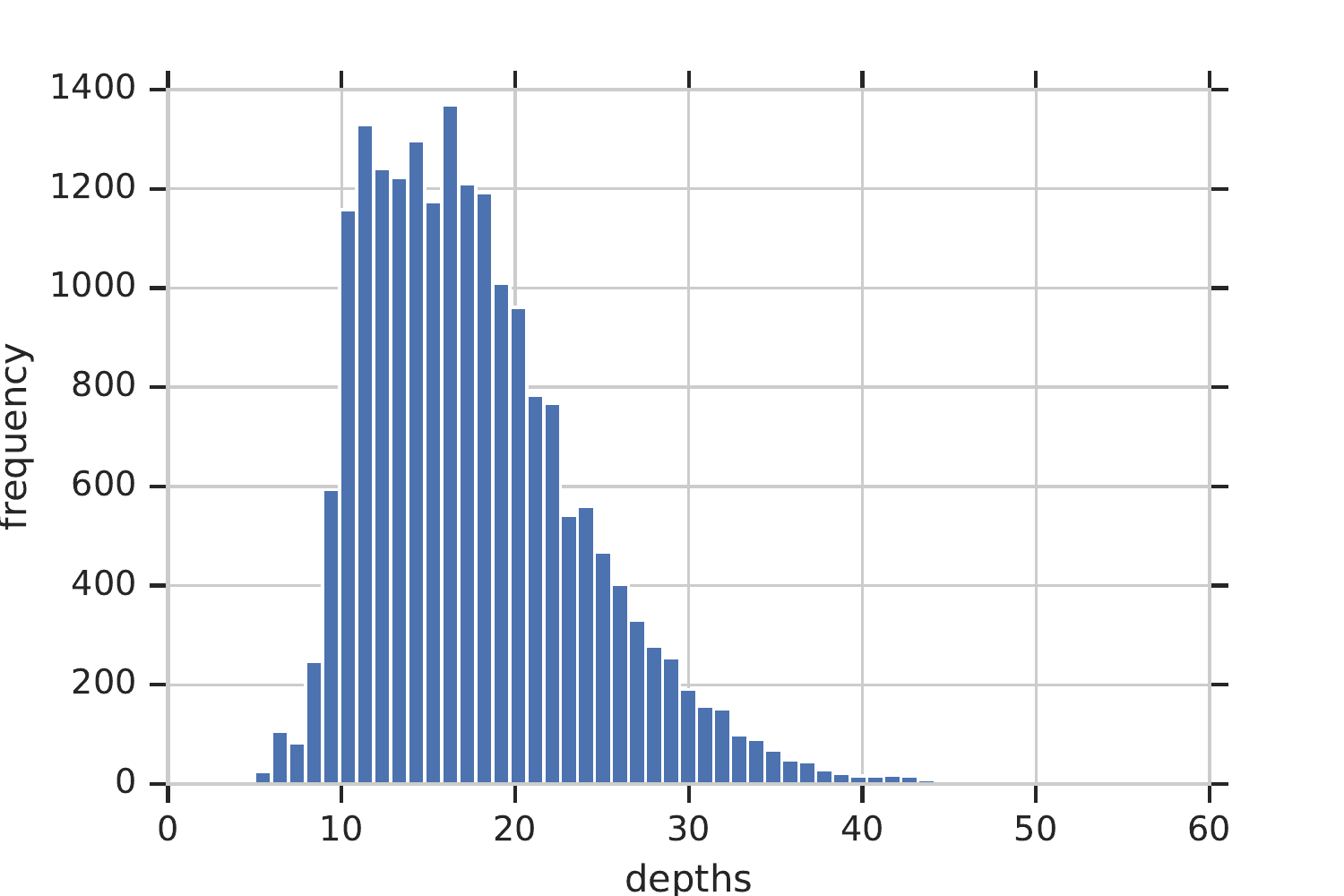}